\documentclass[10pt,journal,compsoc]{IEEEtran}

\usepackage{cite}

\usepackage{booktabs} 
\usepackage{bbm}
\usepackage{wrapfig}
\usepackage[export]{adjustbox}
\usepackage{amsmath}
\usepackage[ruled]{algorithm2e} 

\graphicspath{{figures/}}
\newcommand{\etal}{et al.}

\usepackage{color}

\definecolor{turquoise}{cmyk}{0.65,0,0.1,0.1}
\definecolor{purple}{rgb}{0.65,0,0.65}
\definecolor{dark_green}{rgb}{0, 0.5, 0}
\definecolor{orange}{rgb}{0.8, 0.6, 0.2}
\definecolor{red}{rgb}{0.8, 0.2, 0.2}
\definecolor{brown}{rgb}{0.5, 0.16, 0.16}

\begin{document}
\title{A Non-linear Differential CNN-Rendering Module for 3D Data Enhancement}

\author{Yonatan Svirsky, Andrei Sharf\\Ben Gurion University}

\IEEEtitleabstractindextext{%
\begin{abstract}
In this work we introduce a differential rendering module which allows neural networks to efficiently process cluttered data.
The module is composed of continuous piecewise differentiable functions defined as a sensor array of cells embedded in 3D space.
Our module is learnable and can be easily integrated into neural networks allowing to optimize data rendering towards specific learning tasks using gradient based methods in an end-to-end fashion.
Essentially, the module's sensor cells are allowed to transform independently and locally focus and sense different parts of the 3D data.
Thus, through their optimization process, cells learn to focus on important parts of the data, bypassing occlusions, clutter and noise.
Since sensor cells originally lie on a grid, this equals to a highly non-linear rendering of the scene into a 2D image.
Our module performs especially well in presence of clutter and occlusions. %
Similarly, it deals well with non-linear deformations and improves classification accuracy through proper rendering of the data.
In our experiments, we apply our module to demonstrate efficient localization and classification tasks in cluttered data both 2D and 3D.

\end{abstract}


}

\maketitle

\IEEEdisplaynontitleabstractindextext

\begin{figure*}
	\centering
	\includegraphics{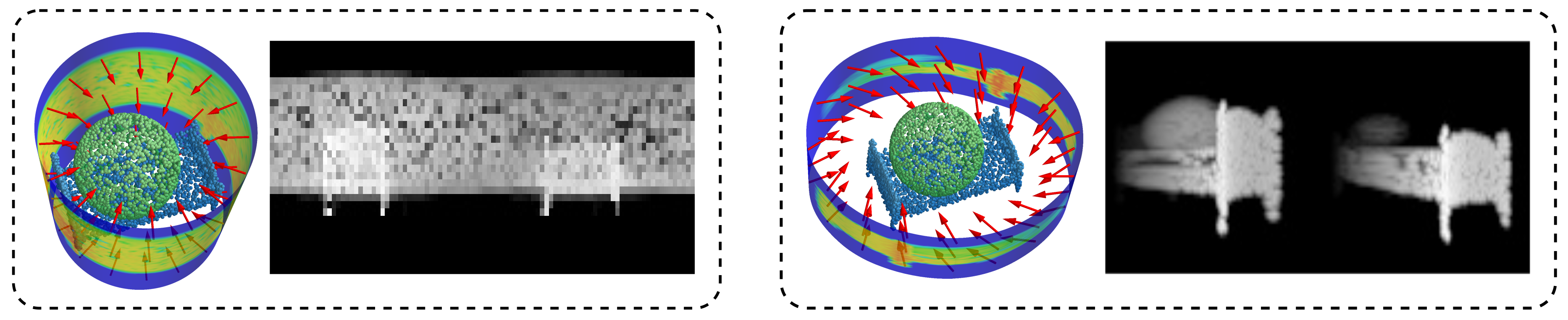}
	\caption{3D clutter suppression via adaptive panoramic non-linear rendering. Left-to-right are the initial scene consisting of a 3D bed and a large spherical occluder and its rendering (mid-left). Our non-linear rendering module transforms sensor cells (indicated by arrows, mid-right), resulting in significant suppression of the 3D clutter (rightmost).}
	\label{fig:panoramic}
\end{figure*}

\IEEEraisesectionheading{\section{Introduction}\label{sec:intro}}

3D shapes, typically represented by their boundary surface, lack an underlying dense regular grid representation such as 2D images.
Optimal adaptation of CNNs to 3D data is a challenging task due to data irregularity and sparseness. Therefore, intermediate representations are typically utilized to allow efficient processing of 3D data with CNNs. In this work we consider a novel non-linear and differential data representation for CNN processing of cluttered 3D point sets.

Alternative data representations have been investigated to allow efficient CNN processing of 3D data~\cite{wu20153d,DBLP:journals/corr/ChoyXGCS16}.
Approaches involve converting the shape into regular volumetric representations, thus enabling convolutions in 3D space ~\cite{wu20153d}.
As volume size is $O(n^3)$ this causes a significant increase in processing times and limiting 3D shapes to low resolution representations.
A common approach is to project the 3D shape into regular 2D images from arbitrary views~\cite{su2015multi}. Nonetheless, selecting optimal views is a challenging task which has been only recently addressed~\cite{Rov18b}.
Nevertheless, in the case of clutter, occlusions, complex geometries and topologies, global rigid projections may lack the expressive power and yield sub-optimal representations of the 3D scene possibly missing important features.

In this paper we introduce a novel non-linear rendering module for cluttered 3D data processing with CNNs.
Our work is inspired by spatial transformer networks~\cite{jaderberg2015spatial} which involve  localization and generation for 2D images. Essentially, our differential rendering module follow this path and suggest an extension of ST networks to differential non-linear rendering in 3D.
Our rendering module is defined as a senor array of cells that sense, i.e. capture the scene through local rendering. Thus, sensor cells are formulated as differential rendering functions with local support that operate on the scene.

Similar to spatial transformer networks~\cite{jaderberg2015spatial}, our module parameters are learnable and therefore it allows the optimization of data rendering towards specific learning tasks in an end-to-end fashion.
We embed our module into various network structures and demonstrate its effectiveness for 3D learning tasks such as classification, localization and generative processes. We focus on clutter, noise and deteriorated data where our differential module makes a clear advantage.

The overall sensor cells array yields a highly non-linear rendering of the scene focusing on features and bypassing noise and occlusions (see figure~\ref{fig:panoramic} for an example).
Our module defines how data is viewed and fed to the neural network. Its differential properties allow to easily plug it into neural networks as an additional layer and become integral to the network optimization. Therefore, we can easily couple between rendering optimization, scene and shape learning tasks.

Cell parameters are optimized during network optimization, thus training to focus on distinctive shape features which adhere to the global learning task.
This is a key-feature to our method as network optimization also optimizes the cells parameters. It allows sensor cells to independently move and zoom in 3D space in order to capture important shape features while avoiding irrelevant parts, clutter and occlusions.
Hence, the rendering module yields enhanced views of the 3D scene that overcome clutter and obtain higher accuracy rates in comparison with other 3D-based approaches.
In our results we demonstrate its utilization to challenging 3D point learning tasks such as classification, localization, pose estimation and rectification. We show applications of our module to several network topologies and supervision tasks.

To summarize, our differential rendering module makes the following contributions:
\begin{itemize}
\item A novel selective non-linear 3D data viewing approach that adheres CNN learning tasks
\item 3D data enhancement through feature attenuation and clutter suppression 
\item The rendering module is differential and thus easily fed into CNNs enabling challenging 3D geometric tasks.
\end{itemize}

\section{Related Work}
\label{sec:related}

Shapes, typically represented by their boundary surfaces are essentially sparse structures embedded in 3D. Applying CNNs in 3D space is not straightforward due to the lack of a regular underlying domain.  In the following we discuss different CNNs methods for 3D data processing.

A long standing question regarding 3D shapes is whether they should be represented with descriptors operating on their native 3D formats, such as voxel grids or polygonal meshes, or should they be represented using descriptors operating on regular views.

Several attempts have been made to represent 3D shapes as regular occupancy 3D voxel grids ~\cite{wu20153d,Maturana2015VoxNetA3}.
To improve the efficiency of volumetric structures, octrees were suggested ~\cite{DBLP:journals/corr/RieglerUG16,Wang:2017:OOC:3072959.3073608}. 3D shape volume sparsity is handled by defining convolutions only on the occupied octree cells.
To handle the sparsity of 3D inputs, a novel  sparse
convolutional layer which weighs the elements of the convolution
kernel according to the validity of the input pixels has been presented~\cite{DBLP:journals/corr/abs-1708-06500}. Their  sparse convolution layer explicitly considers the location of missing data during the convolution operation.
In general, volumetric structures are inefficient as 3D shapes are typically sparse, occupying a small fraction of voxels.  They  have a limited resolution and thus cannot represent fine geometric 3D details. To address these limitations, complex structures and convolutions are required.

Multiple 2D views representations of 3D shapes for CNN processing have been previously  introduced~\cite{su2015multi}. Authors introduce a novel CNN architecture that combines information from multiple 2D views of a 3D shape into a single and descriptor offering improved recognition performance. To deal with invariance to rotation, they perform a symmetric max-pooling operation over the feature vectors.
Similarly, DeepPano~\cite{shi2015deeppano} use a cylindrical projection of the 3D shape around their principle axis. They use a row-wise max-pooling layer to allow for 3D shape rotation invariance around their principle axis.
Spherical parameterization were also suggested in order to convert 3d shapes surface into flat and regular `geometry images'~\cite{sinha2016deep}. The geometry images can then be directly used by standard CNNs to learn 3D shapes.
Qi \etal \cite{Qi2016VolumetricAM} present an extensive comparative study of state-of-the-art volumetric and multi-view CNNs.
2D projection representations are sensitive to object pose and affine transformations. 2D views typically yield partial representations of the 3D objects with missing parts due to self occlusions and complex configurations.

Mesh representations have been processed with CNNs performing in spectral domain ~\cite{masci2015geodesic,boscaini2015learning}. A local frequency analysis is utilized for various learning tasks on meshes. Nevertheless, these methods are currently constrained to valid manifold meshes only.
Similarly, a gradient based approach for mesh rasterization which allows its integration into neural networks was presented~\cite{kato2018renderer}. Nevertheless, their focus is on the mapping between 3D mesh faces and the 2D image space, i.e. the raster operation. This allows learning various mappings between 2D and 3D allowing to perform texture transfer, relighting and etc.
Our method has a different focus and aims at differential non-linear rendering of 3D geometry for improving learning tasks.

3D shapes represented by point-sets have been previously introduces for CNN processing.
PointNet~\cite{DBLP:journals/corr/QiSMG16} introduces a novel neural network that performs directly on the 3D point representation of shape. To account for point permutation invariance, authors  calculate per point high dimensional features followed by a global max pooling. Convolution is possibly replaced with permutation invariant functions to allow processing unordered point-sets~\cite{ravanbakhsh2016deep}, .
In a followup, PointNet++~\cite{DBLP:journals/corr/QiYSG17} enhances PointNet with the ability to recognize fine-grained patterns and complex structures. Authors apply  PointNet recursively and define a hierarchical neural network capable to learn local features with increasing contextual scales.
Several followups define local point-based descriptors and representations for efficient 3D point cloud analysis~\cite{Huang2017,ben18,matan18}.
To tackle the locality of PointNet, the spatial context of a point is considered~\cite{3dsemseg_ICCVW17}. Thus, neighborhood information is incorporated  by processing data at multiple scales and multiple regions together.
These works attempt to define CNN point processing as permutation invariant. Ours sensor cells adapt and focus to the shape surface and efficiently sidestep noise, outliers and occlusions.

\begin{figure}[t]
	\centering
	\includegraphics[width=0.5\linewidth]{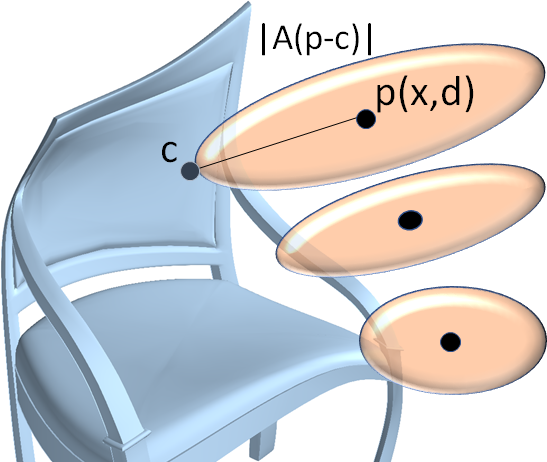}
	\caption{Rendering functions may be formulated as radial kernels applied to the Euclidean distance norm between the sensor cells $p$ and the shape points $c$ after an affine transformation $A$. Kernel parameters may be independently parameterized, yielding nonlinear 3D views and transformations of the  shape.}
	\label{fig:modul}
\end{figure}

Similar to us, ProbNet~\cite{NIPS2016_6416} use a sparse set of probes in 3D space and a volume distance field the sense shape geometry features. Probe locations are coupled with the learning algorithm and thus are optimized and adaptively distributed in 3D space. This allows to sense distinctive 3D shape features in an efficient sparse manner.
Our work differentiates from this work by introducing a differential rendering module that is defined analytically and its derivation is closed form. Thus, our cells move and focus on salient parts of the object bypassing occlusion and noise.
Cells lie inherently on a regular grid and therefore can be efficiently processed by convolutional layers while irregularity of their probes does not lend itself naturally to CNNs leading to ad-hoc and shallow net structures.

Spatial transformers~\cite{jaderberg2015spatial}  introduce a 2D learnable module which allows the spatial manipulation of data within the network.
GAN is used in conjuncture with Spatial Transformer Networks to generate 2D transformations in image space ~\cite{WangErsin2018}. Their goal is to generate realistic image compositions by implicitly learning their transformations.
Our work is inspired by spatial transformer networks extending it into 3D. Due to the nature of 3D space, we use multiple differential cells which move and deform to sidestep occluders and noise and focus on the 3D object.

Recently, 3D point clouds were transformed into a set of meaningful 2D depth images which can be classified using image classification CNNs~\cite{Rov18b}. Similar to us, they define a differentiable view rendering module which is learned within the general network. Nevertheless, their module defines global rigid projections while ours consists of multiple cells that train to move independently, avoiding occlusions and clutter and yielding non-linear 3D projections.

\begin{figure*}[t]
	\centering
	\includegraphics[width=0.9\linewidth,trim={0 0.5cm 0 0}]{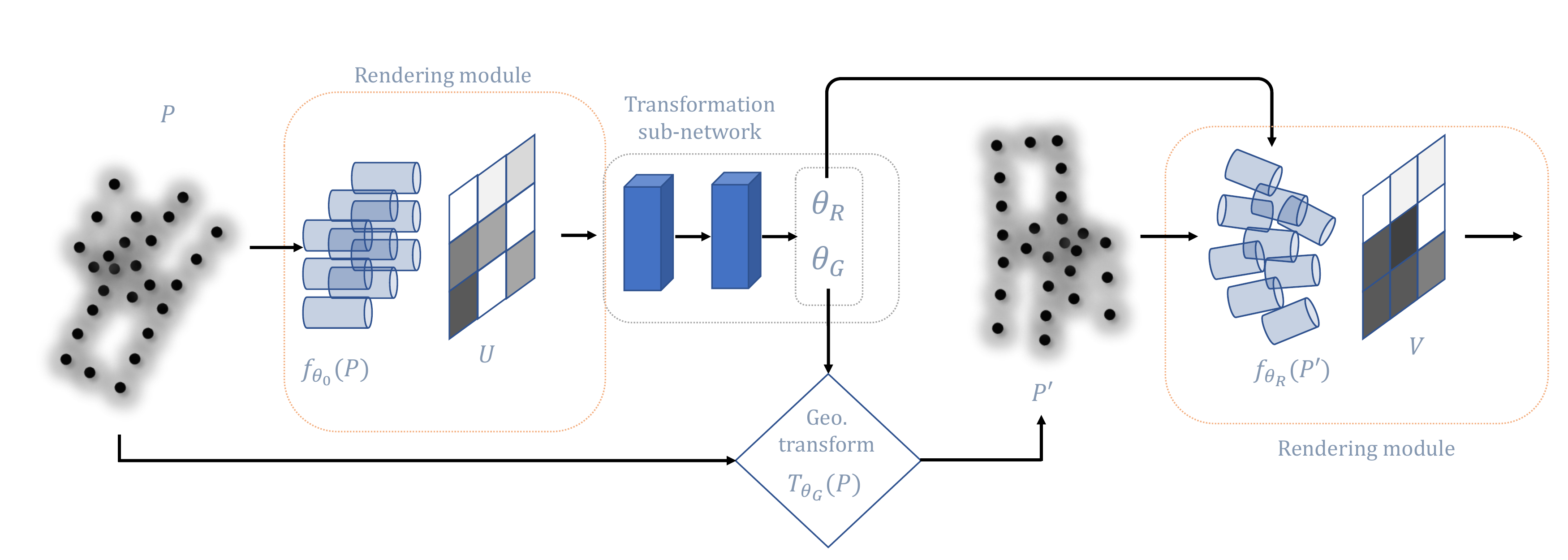}
	\caption{Overview of our rendering module incorporated in neural network. The module takes as input a point cloud $P$ and generates, i.e. renders an image $U$. $U$ is then fed to a sub-network which outputs a set of transformation parameters $[\theta_R, \theta_G]$. $\theta_R$ is applied to the sensors rendering function and  $\theta_G$ is applied to $P$ producing a transformed point cloud $P'$. The final output image $V$ is produced by rendering $P'$ using the new rendering parameters $\theta_R$.}
	\label{fig:pipe}
\end{figure*}


\section{Differential Rendering Module Details}
\label{sec:module}

Our method takes as input a 3D shape represented by a point set sampling of its surface. Essentially, data passes through our rendering module, which produces intermediate 2D renderings. These are then fed to a neural network for various learning tasks.
Our rendering module consists of a grid array of \emph{sensor cells}, defined independently by their viewing parameters and 3D position in space.

Given a 3D point cloud $C$, we define a sensor cell $p$ to have a 3D position in space, a viewing direction and sensing parameters.
The differential rendering cell function is defined as:
\begin{center}
	\begin{align}
		pr(p, C; \phi, \psi) = \phi(\{\psi(r_{p,c})|c \in C\})
	\end{align}
\end{center}
where $p$ is the cell defined by the tuple $(x_p, d_p)$, in which $x_p$ is the 3D location, and $d_p$ is the viewing direction.
$\phi$ is the reduction function and $\psi$ is the kernel function operating on $r_{p,c}$ which is the view transformed distance.

\subsubsection*{View transform.}

We define $r_{p, c} = A_p(x_p-c)$ and $A_p = Scale_p \cdot Rot_p$ is a linear matrix consisting of a rotation $Rot_p$ of the sensor cell direction and an elongation, i.e. scale in view direction, $Scale_p = diag([1,1,s])$. $A_p$ operates on the difference between the cell location and a given point (Figure~\ref{fig:modul}).

If the distance between points in $r_{p, c}$ is defined as a Euclidean norm, then the metric induced by $A_p$, the linear transformation of the norm, can be interpreted as elongating the unit ball of the Euclidean distance metric (Figure ~\ref{fig:modul}).

This is equivalent to replacing the Euclidean distance metric with the Mahalanobis distance metric, defined as:
\begin{center}
\begin{align}
\psi(r_{p, c}) = \sqrt{(x_p-c)^T A^T A (x_p-c)}
\end{align}
\end{center}

The use of Mahalanobis distance can be interpreted as a relaxation of
the range function since:
\begin{center}
\begin{align}
\lim\limits_{s\rightarrow 0} \frac{1}{s} \min\limits_{c\in C}( {\lVert I_s(x_p-c)\rVert})= \hat{c}
\end{align}
\end{center}
where $I_s$ is a diagonal matrix with non-zero entries $(1, 1, s)$, and $\hat{c}$ is the closest point in $C$ intersecting the ray emanating from $x_p$ with direction vector $(0, 0, 1)$ (assuming $C$ is finite or consisting of a 2D smooth manifold).

\subsubsection*{Kernel function.}

$\psi$ is a kernel function that operates locally on the 3D vector data, denoted as $r_{p, c}$. It defines the cell shape and the interaction between the cell $p$ (direction and shape) and points $c \in C$.
Kernel functions take as input the difference between the cell location and the locations of 3D data points. This difference may be defined as a norm scalar field or vector field, yielding different interpretations of the data.

A straightforward approach for defining $\psi$ is as a standard kernel function  $K (\|r_{p,c}\|)$ where $K$ is some kernel (e.g. Gaussian, triangular, etc.) on some distance norm.

It is also possible to define $K(\cdot)$ as a separable function in $XY$ and $Z$, i.e., $K(x,y,z)=f(x,y)\cdot g(z)$.  This allows using a bounded kernel in the ($XY$) image plane weighting an unbounded feature function in depth ($Z$). This considers the projection of points onto the viewing plane and down-weighting points which project far from the cell while being sensitive in the perpendicular direction.

\subsubsection*{Reduction function.}

$\phi$ is a reduction function that maps a set of values to a scalar. In this work we use $max$, and $sum$ for $\phi$ as range and density reduction correspondingly.
Specifically, the range function is defined as:
\begin{center}
\begin{align}
R(s) = \max\limits_{c \in C}{\psi(r_{p, c})}
\end{align}
\end{center}
and the density function is defined as:
\begin{center}
\begin{align}
D(s) = \sum\limits_{c \in C}^{}{\psi(r_{p, c})}
\end{align}
\end{center}
Naturally, these are permutation invariant functions that are insensitive to point ordering, hence adequate for robust processing of irregular point clouds.

\subsubsection*{Differentiability.}

Since sensing functions are composed of continuous, piecewise differentiable functions, their composition is also continuous and piecewise differential. Thus, we can calculate the gradient of the sensing function w.r.t its input almost everywhere.

\subsection{Learning with Differential Rendering Modules}
\label{sec:learn}

The regularity of the sensor cell array allows us to easily generate a 2D image corresponding to the current cells views.
In essence, each pixel in the image is associated with a cell in 3D space, and its value is the output of its corresponding sensing function.

Within this framework, both sensor cells and the input data itself can undergo transformations.
To learn them, we use the differential rendering module in conjunction with a spatial transformation network. The basic process is as follows: the render module is applied to the input shape, producing a 2D image. This image is then fed to a CNN-based transformation sub-network, which produces new rendering parameters $\theta_R$, and geometric transformation parameters $\theta_G$. A new shape is then obtained by applying function $T$, parameterized by $\theta_G$, to the input point cloud. The result is rendered again, using $\theta_R$ (Figure~\ref{fig:pipe}).

\subsection{Clutter Suppression and Attenuation}
We consider the case where 3D scenes contain significant clutter and occlusions which interfere with shape learning tasks. Note that in such cases where the shape is highly occluded, simple affine transformations of the scene are not sufficient to recover a clean view.

Hence, our objective here is twofold:
\begin{enumerate}
\item suppress clutter and other irrelevant information in the scene
\item augment informative features that cannot be seen with standard projections
\end{enumerate}

We achieve this by enhancing the kernel function to have variable sensitivity in its Z direction.
Specifically, the attenuation function for cell $p$ is defined as:
\begin{align}
	\omega^{(p)}(z)= 1 - h\big(\chi^{(p)}(z\big))
\end{align}

where $\chi^{(p)}$ is the Gaussian mixture model for $p$:
\begin{align}
\chi^{(p)}(z) =  \sum_{i=1}^{n} a_i^{(p)} * \exp\bigg(-\Big(\frac{z - c_i^{(p)}}{\sigma_i^{(p)}}\Big)^2\bigg)
\end{align}

and $h$ is either $tanh$ or the Softsign function:
\begin{align}
Softsign(x) =  \frac{x}{1+\lvert x\rvert}
\end{align}

To apply the above suppression mechanism to our rendering pipeline
we modify the kernel function  to
incorporate the cell dependent attenuation field:
\begin{align}
K^{(p)}(x,y,z)=f(x,y) \cdot g(z) \cdot \omega^{(p)}(z)
\end{align}
Intuitively, points with small attenuation coefficients effectively get "pushed back", and thus are less likely
to contribute to the output of a depth rendering function.

Parameters of the attenuation functions are learned along with the other transformation parameters by the transformation network.

\subsection{Learning Iterative 3D Transformations}

We can easily plug our differential rendering module into neural networks that perform general 3D shape learning tasks. This enables transformation network to implicitly learn 3D shape transformations that assist with the core learning problem.

Here, our differential module is used in conjunction with geometry optimization tasks such as shape rectification and pose alignment.
Such tasks, involve fine-grained geometric transformation computations which may be achieved through an iterative transformation scheme utilizing our differential rendering module.
Multi-step and iterative approaches have shown to be efficient in computing coarse estimates in initial steps that are corrected in later steps~\cite{Carreira_2016_CVPR,toshev2014deeppose,Singh2016litland}.
In our case, throughout the optimization process, sensor cells generate various 2D renderings of the scene while a geometric prediction network computes transformation parameters that are applied to the 3D shape through a geometric transformation module iteratively.

A key idea in utilizing our rendering module in order to apply iterative transformations is that they are directly applied on the geometry with no loss of information.
In contrast, in methods that operate on the geometry through an intermediate medium (such as a volumetric grid), applying successive non-composable transformations requires multiple interpolation steps, which results in loss of information.

It is also possible to apply the rendering module as part of a Generative Adversarial Network (GAN) without direct supervision. In this setting, the transformation network, together with the rendering module, act as the generator, which apply transformations to the shape using an adversarial loss. The discriminator takes as input transformed shapes produced by generator, along with a set of samples drawn from the ground truth distribution.

\begin{figure}[t]
	\centering
	\includegraphics[width=\linewidth]{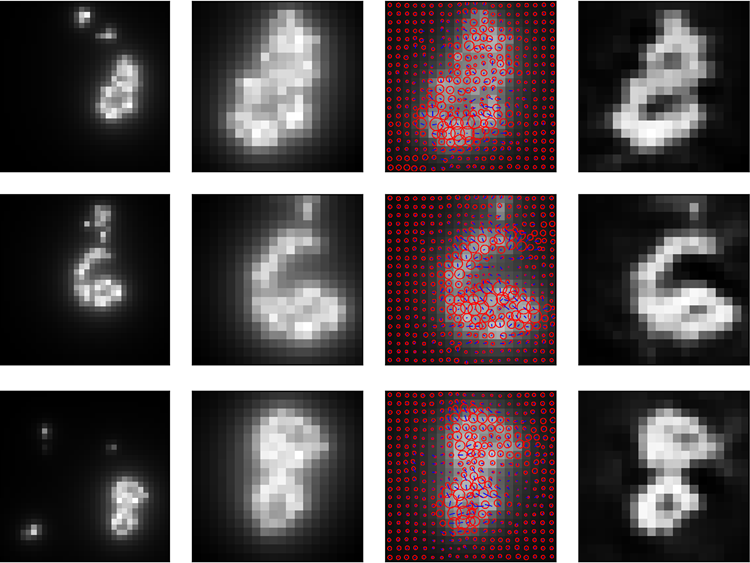}
	\caption{Feature enhancement on cluttered MNIST (rows depicting three different results). Columns, left-to-right are the sampled digits with clutter as initially viewed by the cell grid, localization of digits, cells overlayed (circles indicating magnitude of kernel bandwidth, blue arrows indicating cell spatial shift).  Rightmost is a featured enhanced (i.e. sharpened) point cloud generated by the cell kernels.}
	\label{fig:mnist}
\end{figure}

\begin{table}[h]
	\centering
	\caption{Cluttered-MNIST Classification}
	\label{tab:mnist-table}
	\begin{tabular}{ll}
		Transform                       & Accuracy                    \\ \hline
		\multicolumn{1}{|l|}{None}      & \multicolumn{1}{l|}{78.36\%} \\ \hline
		\multicolumn{1}{|l|}{XY,Scale}  & \multicolumn{1}{l|}{89.65\%} \\ \hline
		\multicolumn{1}{|l|}{Per-Cell} & \multicolumn{1}{l|}{90.87\%} \\ \hline
	\end{tabular}
\end{table}

\subsection{Efficient Render Module Implementation}

Cell functions are at the core of our network optimization. Specifically, we optimize the cell viewing parameters generating new renderings of the input 3D scene. This requires multiple evaluations of the cell functions and their gradients.
Potentially, evaluation of the cell function requires computation of all pairwise interactions, i.e. point differences, between the cell and input point cloud.

To this end, we define an efficient implementation of the cell function computation. We use a KD-tree spatial data structure and approximate cell and point cloud interaction using efficient approximations.
Thus, we define efficient queries between the cell kernel and the KD-tree structure.

Cell parameters, i.e. cell location, direction and kernel function, define the cell interaction with the point cloud. This yields specific queries into the point cloud. Since the cells kernel functions are support-bounded, we can incorporate the KD-tree query inside the neural network, while still maintaining function continuity


Each kernel function that we use has a bounded support.
In the case of using Mahalanobis distances, ellipsoid kernels may be approximated using oriented bounding boxes in the direction of the ellipsoid.
Similarly, in the case of separable cell functions, the query reduces to an infinite height cylinder whose base is the $XY$ viewing plane and height is the $Z$ direction. Nevertheless, since the 3D scene is in itself bounded, the cylinder may also be efficiently approximated by an oriented bounding box in $Z$ direction.
In both cases, querying the KD-tree with oriented bounding boxes is efficient and equals to performing intersection tests between boxes and the KD-Tree nodes' boxes at each level of the tree .

An additional performance improvement may be achieved in the case where cells positional parameter is constrained to a regular grid on the plane. This equals to cells positions undergoing a similarity transformation (note that other viewing parameters such as direction and kernel shape remain unconstrained).

In this case, there exists a one-to-one mapping between the 3D point cloud and the cell grid on the image plane defined by a simple orthographic projection. Thus, we define a binning from 3D points in space into the grid cells corresponding to the sensor cells. This yield a simple query of the 3D points that is of linear time in the point size (Algorithm~\ref{alg:hash}).

\begin{algorithm}[t]
	\KwData{3D point cloud $C$, $m\times n$ cell grid,$\phi$, $\psi$}
	\KwResult{$m\times n$ rendered image}
	initialization $m\times n$ zeros array $A$\;
	\For{$c \in C$}{
		$i,j = \lfloor c_x \rfloor, \lfloor  c_y \rfloor\ $\;
		\For{$(k,l) \in Neighborhood (i,j)$}{
			$A[k,l] = \phi [\{A[k,l],\psi(c-[k,l,0]^T) \}]$\;			
		}
	}
	\caption{Orthographic binning}
	\label{alg:hash}
\end{algorithm}

\begin{figure*}[]
	\centering
	\includegraphics[width=0.91\linewidth]{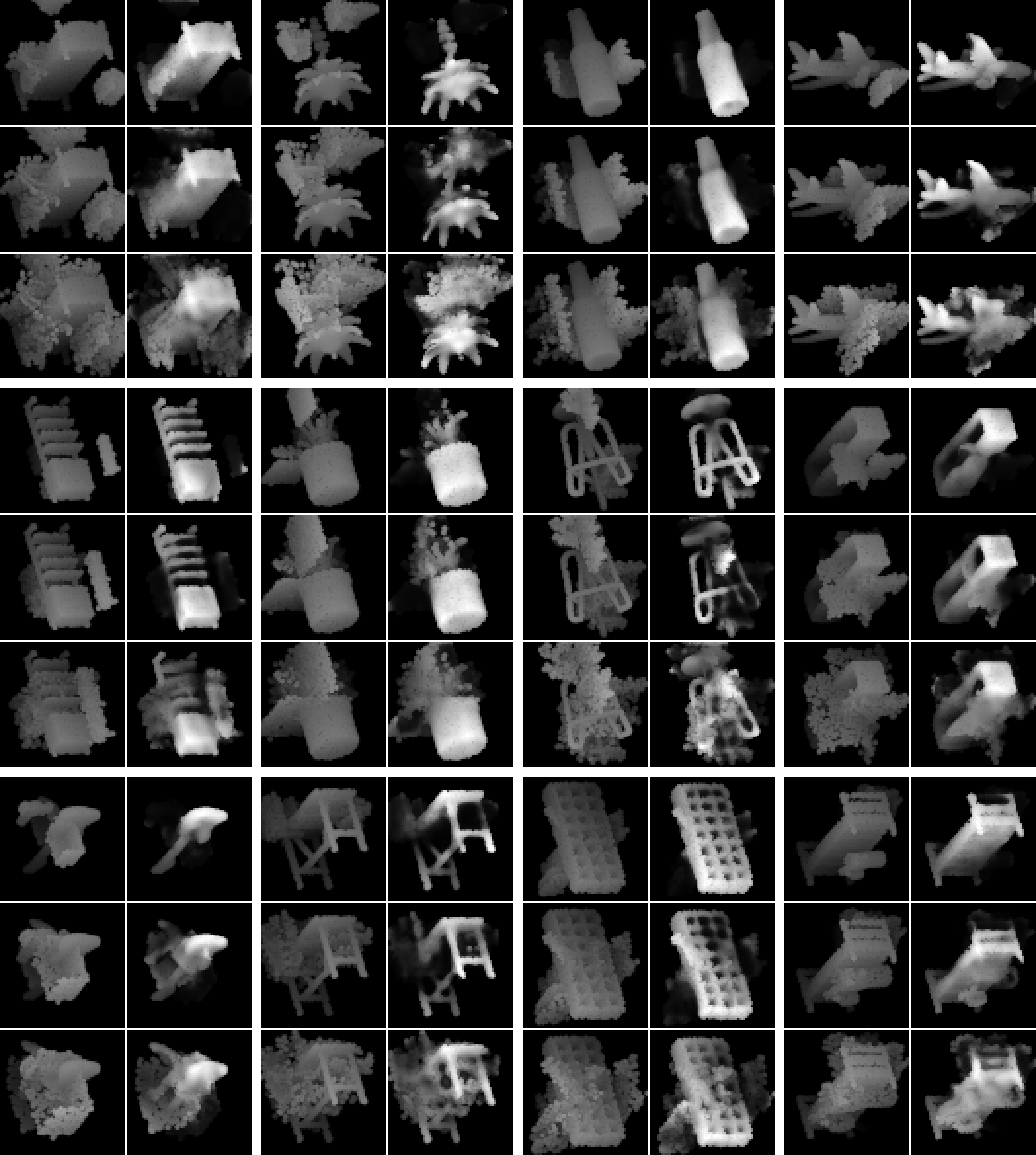}
	\caption{3D clutter suppression results on 12 objects  with 3 different clutter scales. Column-pairs depict before-and-after clutter suppression for each object. Rows-triplets depict the 3 clutter scales per object.}
	\label{fig:clutter_scale_before_after}
\end{figure*}

\section{Experiments and Results}
\label{sec:results}

We have tested the effectiveness of our differential rendering module on various datasets with different network structures. Experiments consists of 3D transformations and renderings for various shape tasks that were implicitly learned in weakly supervised scenarios. In each of the experiments, the structure of the models we use is similar: The input is initially rendered using sensor with constant parameters, or global parameters which are not conditioned on the input;
To describe the models used, we adopt the notation C[n,k,s] for convolution with n filters of size (k,k), with stride s, (A$|$M)P[s] for (average$|$max)-pooling with window size and stride s, FC[n] for fully-connected layer with n output units, LSTM[m] for LSTM~\cite{hochreiter1997long} RNN cell with m hidden units.

\subsection{Non-linear Localization from Cluttered-MNIST}

The point-cloud sampling of MNIST~\cite{Lecun98} is generated by the same protocol as in \cite{jaderberg2015spatial}. More specifically, we use the dataset provided by \cite{lasagne}.
For each image sample, 200 points are sampled with probability proportional to intensity values.
Thus, our MNIST dataset consists of 2D point-clouds, setting their z coordinate as for all sensor cells to 0.
Cluttered-MNIST is based on the original MNIST dataset where original images are randomly transformed and placed into a $2\times$ larger image and clutter is introduced by randomly inserting $4-6$ sub parts of other shapes into the image.

The model consists of an initial rendering, two transformation + rendering phases, followed by classification network. The initial rendering is done using a $40\times 40$ sensor grid, and the dimensions of subsequent generated images are $20 \times 20$. we use density rendering with kernel $K(x)=\frac{1}{1+(\frac{x}{\alpha})^2}$.
The first transformation network is in charge of scaling and translating the input. Its structure is: (C[20,5,1], MP[2], C[20,5,1], FC[50], FC[3]]).
In second phase, the network  generates a $20\times 20\times 3$ parameter array which defines for each of the sensor cells an in plane shift $(\Delta x,\Delta y)$ and bandwidth $\alpha$.
The final classifier has the structure: (C[32,5,1], MP[2], C[32,5,1], MP[2], FC[256], FC[10]).

Figure~\ref{fig:mnist} shows qualitative results of localization of three digits. Using our differential rendering module, the network successfully zooms into the shape cropping out outliers (mid-left col).
The effects of localization on the overall classification accuracy are summarized in table~\ref{tab:mnist-table}.

We compare between classification accuracy with no localization (top row), global translation and scale (mid) and non-linear localization deformations allowing each cell to focus independently (bottom).
We can further utilize the independent cell transformations to perform feature enhancement and sharpening (Figure~\ref{fig:mnist}, rightmost column). This is achieved by considering the cell kernel size as a feature filter, i.e. filtering out unimportant regions in the shape where kernels have a small scale factor.

\begin{figure}
	\centering	\includegraphics[width=\linewidth]{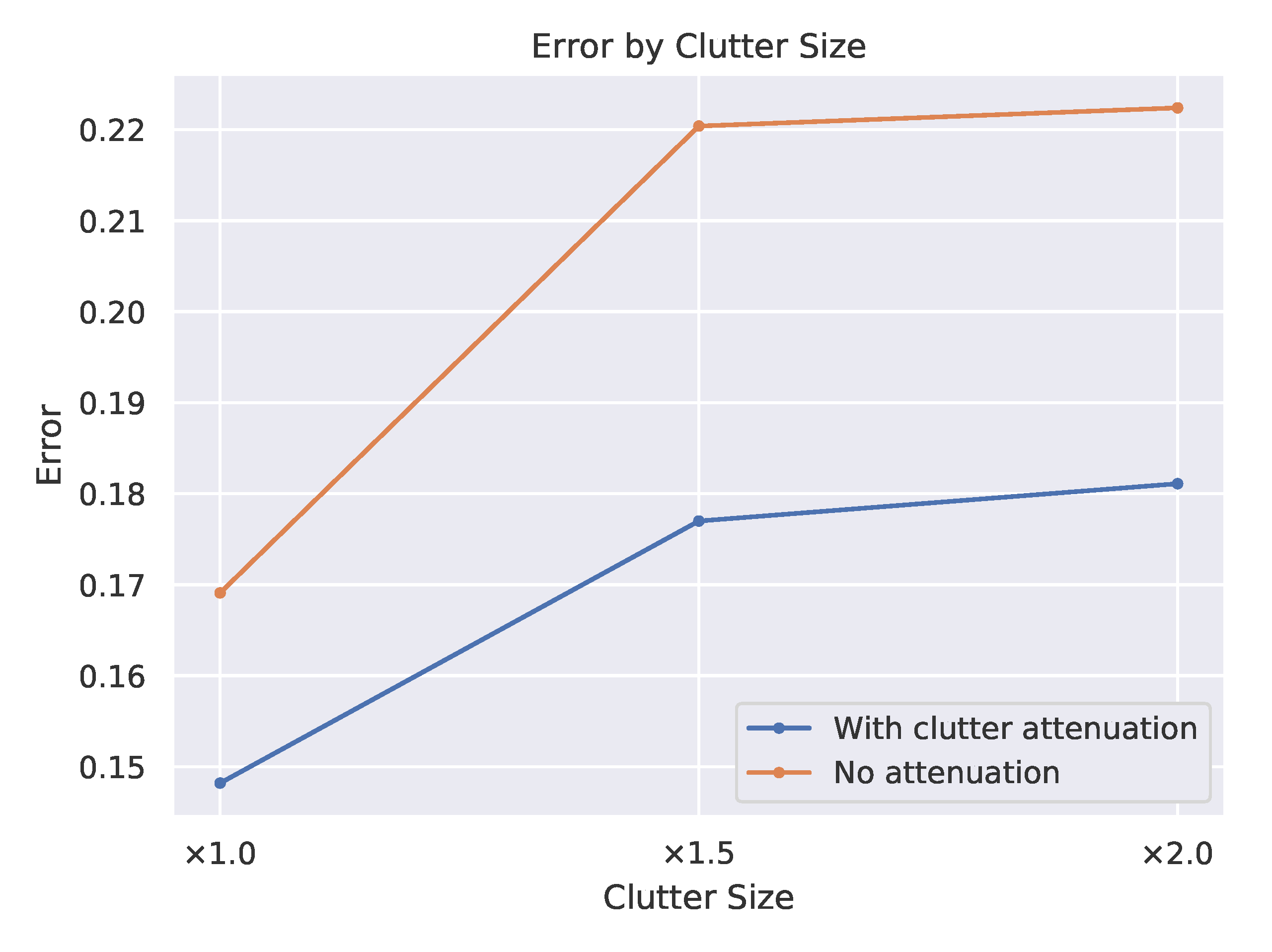}
	\caption{Classification error with and without clutter attenuation, by clutter scale.}
	\label{fig:att_acc_per_scale}
\end{figure}

\begin{figure}
	\centering	\includegraphics[width=\linewidth]{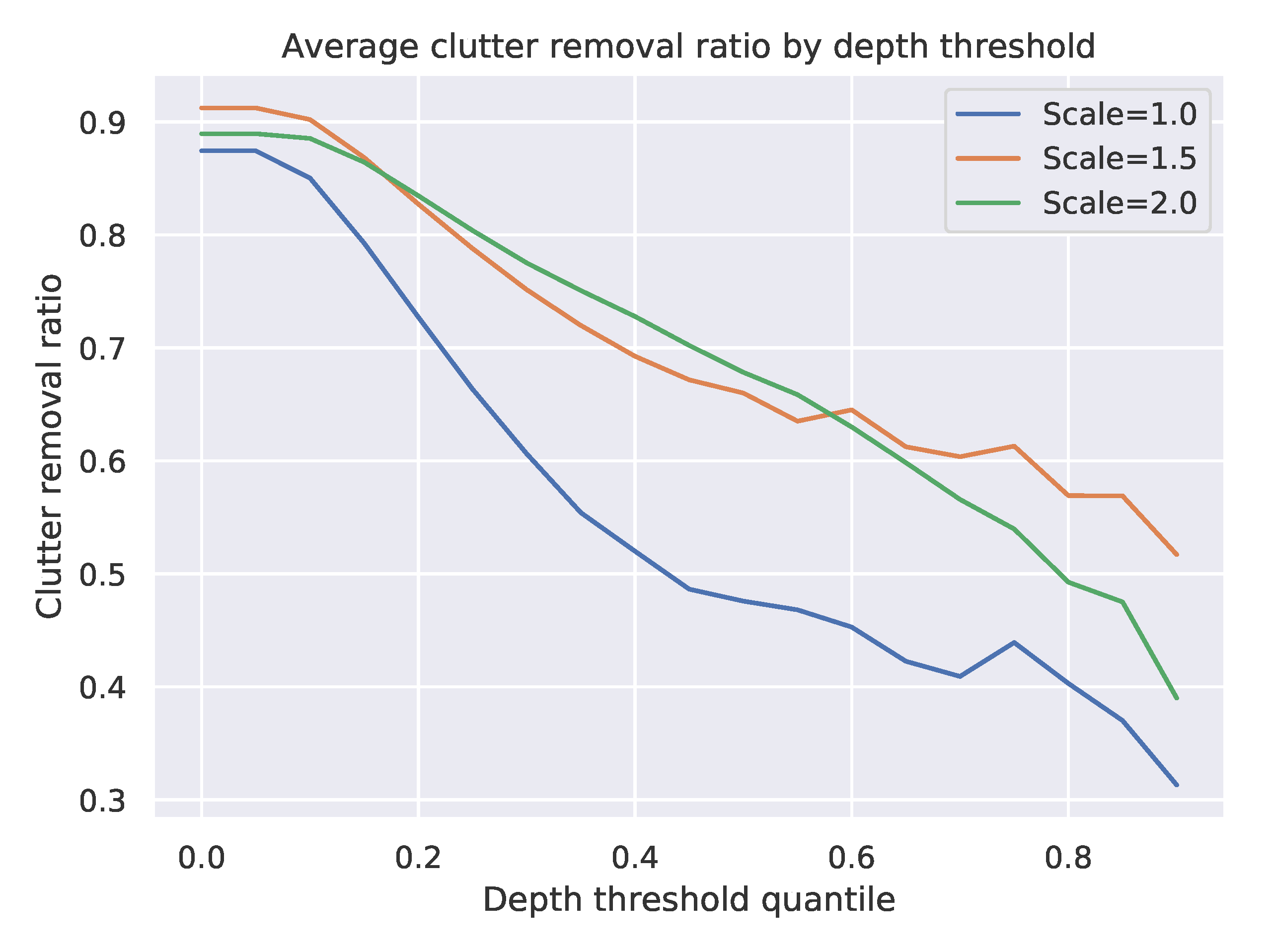}
	\caption{Clutter removal ratio by quantile of depth values.}
	\label{fig:clutter_removal_ratios}
\end{figure}

\begin{table}[]
	\caption{Cluttered-ModelNet40 accuracy}
	\label{tab:clutter_accuracy}
	\centering
	\begin{tabular}{|l|l|}
		\hline
		\multicolumn{1}{|c|}{Method} & \multicolumn{1}{c|}{Accuracy} \\ \hline
		PointNet                     & 69.0 \\ \hline
		Ours (fixed rendering)       & 85.7 \\ \hline
		Ours (adaptive rendering)    & 87.2 \\ \hline
	\end{tabular}
\end{table}
\subsection{Classification of Cluttered Shapes with Non-Linear Rendering and Feature Attenuation}
\begin{figure*}[]
	\centering	\includegraphics[width=\linewidth]{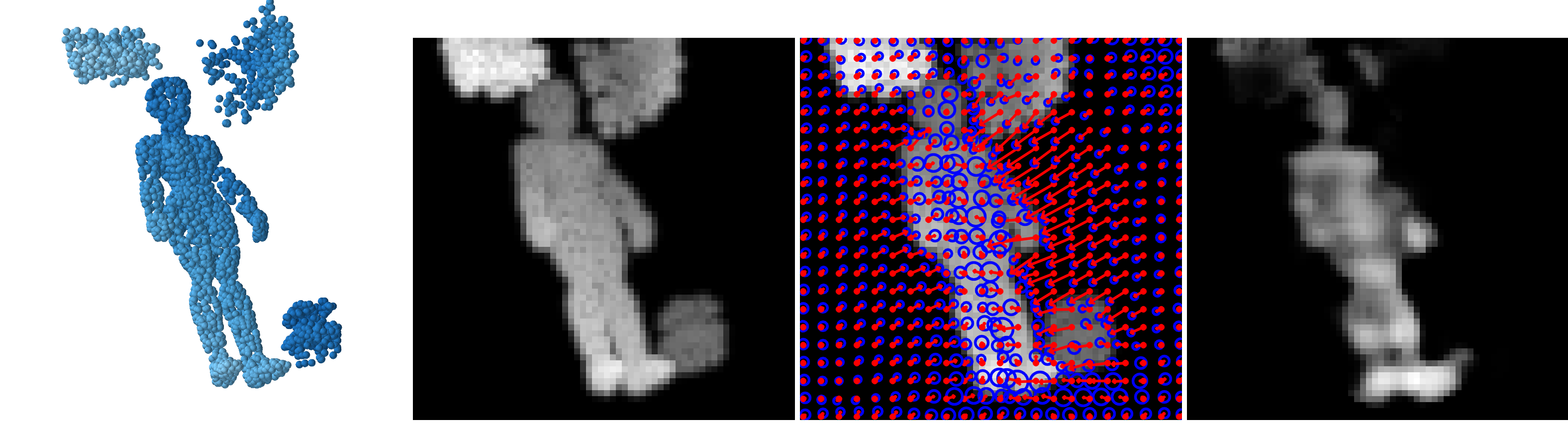}
	\caption{3D shape enhancement and attenuation through non-linear rendering. Left-to-right are the original point cloud, rendered initial point-cloud image, sensor cells transformations and the resulting rendered point cloud.}
	\label{fig:probe_shift}
\end{figure*}
Here we investigate the contribution of our clutter suppression module, coupled with the rendering module, to the classification of objects in cluttered scenes. Specifically, we use a cluttered version of Modelnet40\cite{wu20153d}, generated using a protocol similar to that of cluttered-MNIST: First, we normalize the dataset by uniformly scaling and translating samples to lie inside the unit ball. We then sample points on the meshes uniformly in order to get a point-cloud version of the object. In addition, for each of the train and test datasets, we create a pool of clutter fragments by cropping sub-parts of the objects using a ball with radius 0.3. Samples in the new dataset contain one whole object and several fragments.
In the first experiment we vary the scale of the clutter fragments in the samples and evaluate the effect of clutter size on the classification accuracy. We compare the results to a baseline model without the clutter suppression module (see examples in figure~\ref{fig:clutter_scale_before_after}). Results are given in figure \ref{fig:att_acc_per_scale}. In addition, we measure the ratio of clutter present in the image before and after applying suppression (figure~\ref{fig:clutter_removal_ratios}). Here the output of the sensor is a $64\times64$ depth image, the attenuation functions use a mixture of 3 Gaussians, in combination with the Softplus activation function. The transformation network for this model is based on a U-Net architecture~\cite{ronneberger2015u} where the number of filters in the downward path are [32, 64, 96, 128], and its output has 32 channels. This is followed by a small CNN (C[64,3,1], C[9,3,1]), which outputs the attenuation functions. We use leaky-ReLUs~\cite{maas2013rectifier} as activation functions for all convolution layers in the transformer network, except for the output; average pooling for down-sampling, and bilinear interpolation for up-sampling. The outputs of the
layers inside the U-Net are normalized using batch renormalization~\cite{ioffe2017batch} with renormalization momentum set to 0.9.

In the rendering module,
the lateral weight kernel used is  $k_a(x) = max(0, (1 - x^2)^{1.65})$, scaled to have a support radius of $1/32$. The depth kernel is $k_b(x)=max(0, 1-x)$.

The classifier consists of 3 convolution + max-pooling stages, starting with 16 channels and doubling them with each convolution; a 256 unit fully-connected layer, and the final classification layer.
We use ADAM optimizer~\cite{kingma2014adam}, with learning rate set to 0.0002 for the classifier and 0.0001 for the transformation network.

\begin{figure*}[]
	\centering	\includegraphics[width=0.95\linewidth]{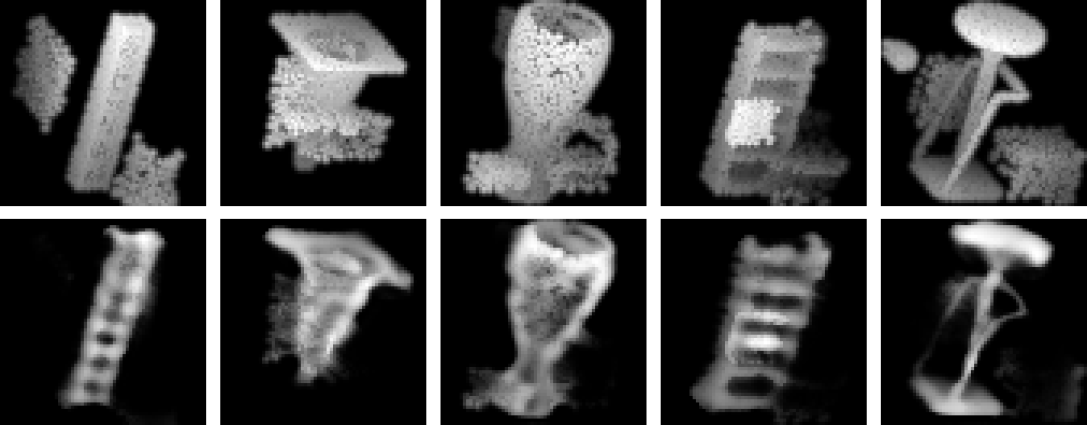}
	\caption{Qualitative evaluation of our 3D clutter removal. Top row are the cluttered objects, bottom row are the clutter-suppressed objects through non-linear sensor cell transformations in 3D space.}
	\label{fig:clutter}
\end{figure*}

We additionally experiment with a similar model, but having a stronger rendering module: In addition to depth, it also renders several density channels that capture more information of the scene; It can translate the
sensor cells along the sensor plane, and vary the sensitivity of each cell to points in its sensing range (for an example of the process, see Fig~\ref{fig:probe_shift}). Here, the
standard density values, which are a function of the lateral distance, are multiplied by kernels that operate on the z values of each point:
$k(x;\mu, \sigma)= \exp(-\frac{\lvert x-\mu\rvert}{\sigma})$
for $\mu$ values \{0, 0.5, 1\}, and $\sigma=0.15$. To these channels we also add the standard density image.
. We apply to all summation based channels the following attenuation function: $f(x) = log(1+\beta\cdot x)$, where $\beta=0.2$.
The lateral kernel used here is $k_a(x) = max(0, (1 - x^2)^\alpha)$, where $\alpha$ is a parameter estimated by the transformation network. Additionally, instead of Softsign, we use $tanh$ for the activation in the attenuation module.  The accuracy of this model is compared against a baseline model without adaptive rendering in Table~\ref{tab:clutter_accuracy}. Figure ~\ref{fig:clutter} shows examples of the depth images produced by the rendering module, before and after transformation.
\subsection{Classification with Adaptive Panoramic Rendering}\label{pano_results}
\begin{figure}[ht]
	\centering	\includegraphics[width=\linewidth]{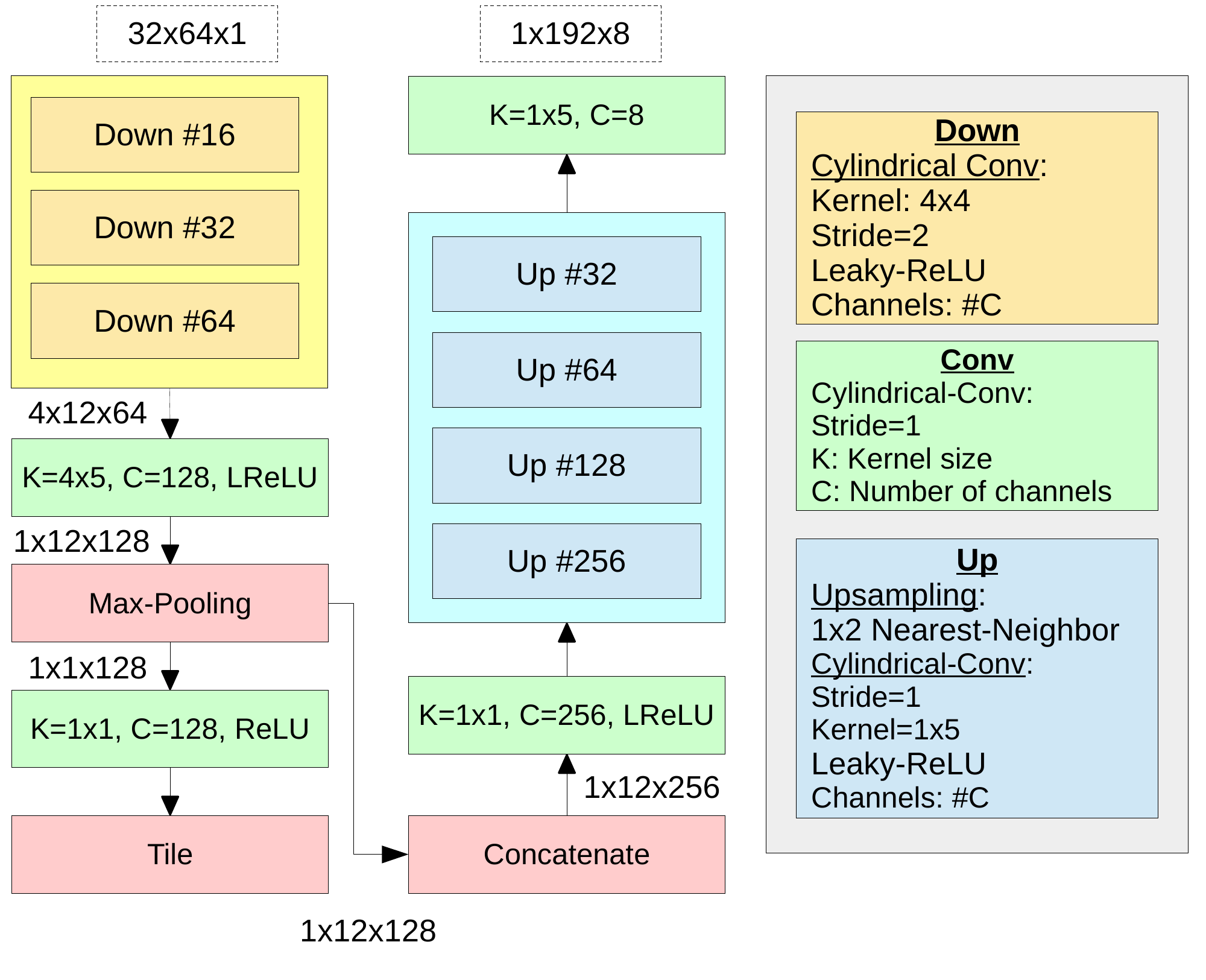}
	\caption{Illustration of transformation network for panoramic rendering.}
	\label{fig:panoramic_loc_net}
\end{figure}

We utilize the rendering module for the computation and optimization of panoramic projections of the 3D scene. Using our rendering module, this can be achieved by placing a cylindrical sensor grid around the scene and representing sensor cells positions in cylindrical coordinates~\cite{shi2015deeppano}.
Such projections are informative and highly efficient in capturing large scale scenes and complex objects.
Specifically, cylindrical panoramic projections allow capturing information of the 3D scene from multiple views at once, as opposed to a single view projection.
Cylindrical representations have also the benefit of being equivariant to rotations around the cylinder's principal axis.
In order to produce a panoramic representation, several modifications are in order: We start by mapping the 2D sensor grid of height $h$ and width $w$ to a cylinder:
\begin{center}
	\begin{align}
		(i, j) \rightarrow \Big(-0.5 \cdot sin(\theta_j),\: \frac{i}{h-1} - 0.5,\: 0.5\cdot cos(\theta_j)\Big)
	\end{align}
\end{center}
where $\theta_j = \frac{j}{w} \cdot \pi$.
The convolution of a cylindrical grid $I$ with height $h$ and circumference $c$, and a 2D kernel $K$ with height $2K_h + 1$ and width $2K_w + 1$, with its middle pixel indexed as (0,0) for convenience, is defined as:
\begin{equation}
\hat{I}(x,y) = \sum\limits_{j=-K_w}^{K_w} \
\sum\limits_{i=-K_h}^{K_h} {I\big((x+j) \,mod\, c, (y+i)\big)K(-j, -i)}
\end{equation}
Initially, the view direction of each cell is towards, and perpendicular to, the Y-axis. The transformation parameters that determine the view source position for each pixel are expressed in the cylindrical coordinate space. More concretely, if pixel $(i,j)$ has been initially mapped to 3D space with the cylindrical coordinates $(\theta_{i,j}, h_{i,j}, r_{i,j})$, after applying the spatial transformation, it would be mapped to
$(\hat{\theta}_{i,j}, \hat{h}_{i,j}, \hat{r}_{i,j})$. In addition to transforming the pixels' sources,
we also allow for changing their viewing direction. The direction is expressed in terms of the difference vector between a destination point and a source point, normalized to have a unit norm. The default position of the $d$ that corresponds to source $s=(\hat{\theta}, \hat{h}, \hat{r})$ (in cylindrical coordinates) is
$(0, \hat{h}, 0)$, (which is the same in either Cartesian, or cylindrical coordinate space).
The transformed destination
point is expressed in terms of a 2D displacement from the default position, along the plane that is perpendicular to the default direction, and coincide with the Y axis.
For the panoramic images, we use a resolution of $96\times32$ for the input image, and $192\times64$ for the output image.
The transformation network structure is described in figure~\ref{fig:panoramic_loc_net}. The final layer generates  8 parameters for each column in the cylindrical grid: angle shift, view shift , and 2 sets of 3 parameters for the top of and the bottom of the column, that determine the radius, height, and vertical orientation of the view. These parameters are linearly interpolated for all other cells in the column.

The network is trained on a modified version of ModelNet40, in which the point cloud consists of both the original sample, together with a sphere of varying size, are placed around the origin (figure~\ref{fig:panoramic}).
We apply Spectral Normalization~\cite{miyato2018spectral} to all layers in the model, except for the output layer of the transformation net, and the final two layers in the classifier. We train the model with using ADAM optimizer with a learning rate of 0.0002. Results are described in table~\ref{tab:panoramic_accuracy}.

\subsection{Iterative Pose Estimation for Shape Classification}
We evaluate the ability of our rendering module to aid in classification of arbitrarily rotated shapes.
Starting with the input point-cloud the network iteratively rotates the shape. Each iteration consists of
rendering the object in its current pose, and feeding the resulting image to a 2D CNN which outputs a new rotation, parameterized by quaternions. This transformation is composed with the previous ones. The resulting rotation is then applied to the input point-cloud,
followed by a rendering of the final rotated shape.
The final image serves as an input to the classifier.
In our implementation, each rendered image has resolution of $80\times80$, with 2 channels - depth and density.

The transformation network, which is the sub-network used to produce quaternions based on the current image, has the following structure: its base consists of 3 convolution layers, with 16, 32, and 64 channels, and kernel size 5x5. The first 2 layers have a max-pool operation, with a down-sampling factor of 2, applied immediately after. The convolution layers are followed by 2 fully connected layers, each with 128 units. The output of the described network is then fed to an LSTM layer with 128 units, which allows to aggregate information over the previous iterations. Lastly, the output of th LSTM cell if fed to a fully-connected layer that produces the quaternion parameters, and initialized to produce the identity transformation. All layers preceding the LSTM layer are followed by a leaky-ReLU~\cite{maas2013rectifier} activation function. The parameters of this networks are shared across all iterations, except for the output layer. the output of the last iteration is fed to the classifier,
which in our case is a ResNet-50 model ~\cite{he2016deep}. We choose to initialize the ResNet model with pre-trained weights, as it has been demonstrated to improve performance and reduce training time~\cite{sharif2014cnn,girshick2014rich}. However, models pre-trained on RGB images require 3-channel inputs, while our input consists of 2 channels, which cannot be semantically mapped to different color channels. Therefore, we adjust the filters of the first convolution layer to work on inputs with 2 by applying a random linear projection from 3 dimensions to 2. In addition to the classifier, we also employ an auxiliary classifier on intermediate images, in order to encourage every iteration to produce good transformations. The auxiliary classifier is a simple fully connected neural network with one hidden layer consisting of 128 units. The inputs to the auxiliary classifier are the latent features produced by the transformation net.
In table~\ref{tab:rot_class} we evaluate our approach in comparison with other point-based methods.
\subsection{Pose Estimation and Shape Rectification with Adversarial Loss}
The data used in the following experiments is based on the point cloud version of Modelnet10~\cite{wu20153d}, where each point-cloud sample is obtained by sampling uniformly the original mesh. Our input data consists of two classes: the original point clouds that are canonically aligned, as well as their geometrically transformed versions.

Instead of the classifier we use a discriminator with a Wasserstein loss function (i.e. defining a Wasserstein GAN~\cite{arjovsky2017wasserstein}).
Specifically, we use the variant referred to as WGAN-LP described in ~\cite{petzka2018on}. This variation of WGAN has been shown to have a more stable optimization and convergence properties.

\begin{figure*}[]
	\centering
	\includegraphics[width=0.9\linewidth]{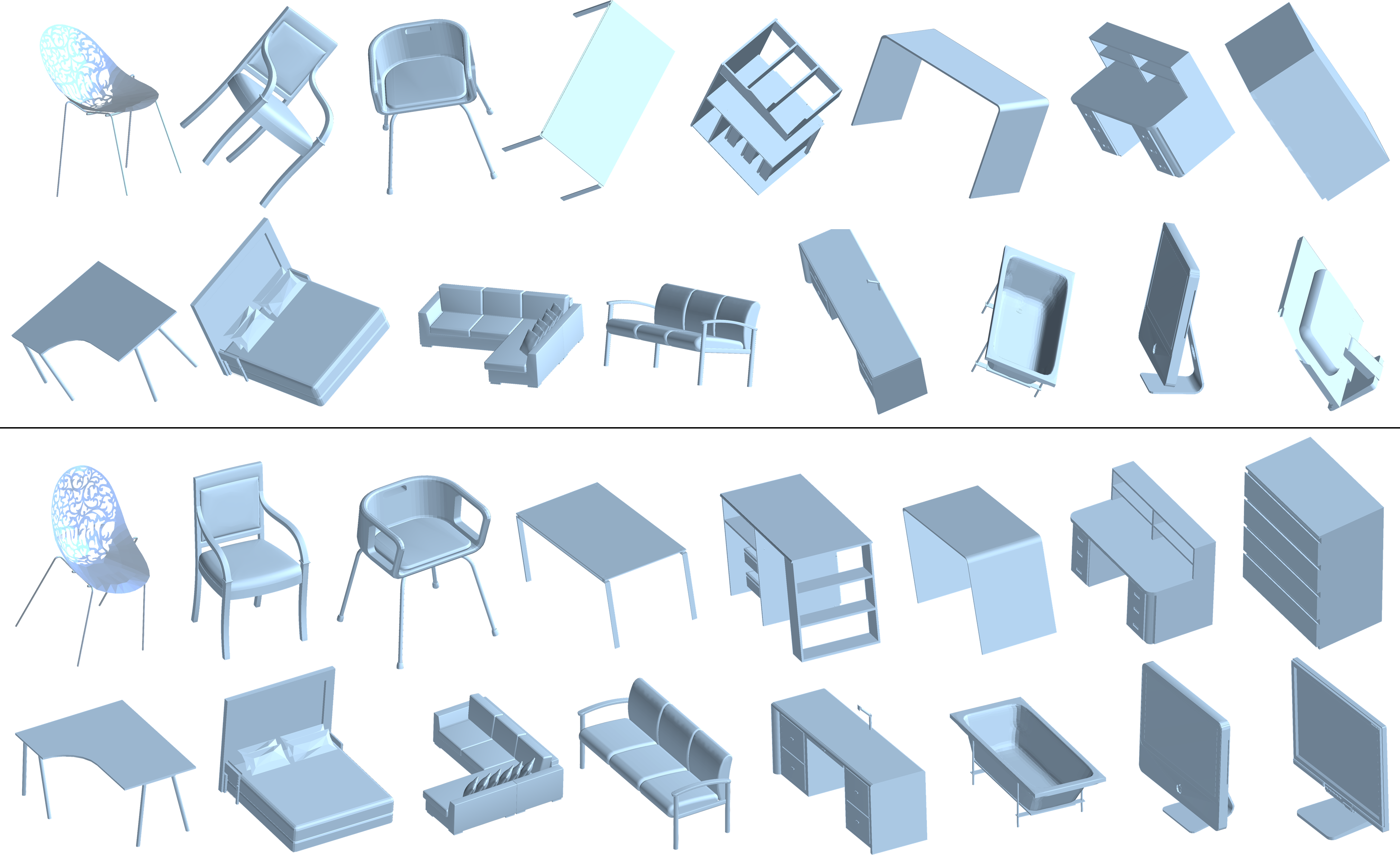}
	\caption{Weakly-supervised 3D pose estimation using our differential rendering module. Given 3D shapes from different classes in arbitrary poses (top), the model implicitly learns a unified canonical pose for all shapes and across classes through a weakly supervised GAN process (bottom).}
	\label{fig:quatgt}
\end{figure*}

\subsubsection{Pose Estimation}
The transformation step consists of 3 steps: Rendering the object in its current pose, computing
the parameters of a quaternion using the transformation network, and applying the rotation represented by
the quaternion to the object.

The transformation network has the structure: (C[20,5,1], AP[2], C[20,5,1], AP[2], FC[128], LSTM[128], FC[4]).
Note that we use here an LSTM cell with 128 units, which potentially allows to aggregate information from rendered views along the iterative process and possibly lead to better precision.
Since rotations are compositional, we aggregate the rotations along the iterative process and apply in each step the aggregated rotation and apply the current rotation on the input.

Figure~\ref{fig:quatgt} shows a gallery of shapes from different classes and with arbitrary orientation (top row). Bottom row demonstrates the alignment of all shapes into a canonical global pose. The model implicitly learns this pose and transforms any shape to it using an adversarial discriminator which operates on a subset of shapes in their canonical pose.
Quantitatively, our method obtains a mean absolute error $err=0.348$ radians in the angle between the ground truth and obtained quaternions. Hence, through the adversarial process, the model learns plausible and meaningful pose estimations that are close to their ground truth.
\subsubsection{Nonlinear shape rectification}
Similar to the above, in order to compute non-linear rectification transformations, The model computes a TPS transformations ~\cite{bookstein1989principal} that rectifies deformed 3D shapes into rectified ones.
 has the structure: (C[32,5,1], AP[2], C[64,5,1], AP[2], FC[128], LSTM[198], FC[32]).
In order to obtain the TPS parameters we interpret the FC[32] output vector as a displacement of a $4\times 4$ grid of control points of the TPS.
Since the control points essentially lie on a 2D plane with restricted displacements, it limits the expressiveness of our model to TPS deformations defined as such. Specifically, 3D points undergo deformations only in the 2D plane of the TPS.
Nevertheless, we found it to be sufficient for rectification of shapes undergoing well behaved non-linear deformations.

\begin{figure}[]
	\centering
	\includegraphics[width=\linewidth]{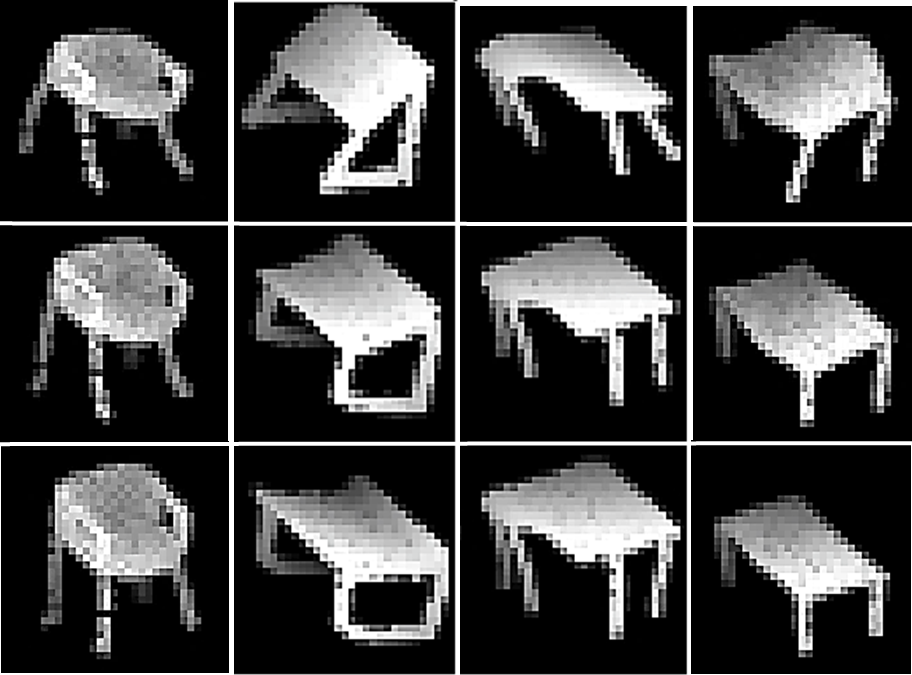}
	\caption{Rendering module optimization iterations  for rectification deformation estimation. Left-to-right are four different shapes which have been deformed using highly non-linear deformations. Rows depict network iterations: starting with the deformed shape(top) and gradually rectifying the shapes (bottom).}
	\label{fig:tps}
\end{figure}

Figure~\ref{fig:tps} shows differential rendering iterations of four different shapes. Starting from deformed shapes, the model implicitly learns the rectification transformation through iterative 3D rendering and transformation of the shape.

\begin{figure*}[]
\centering
\includegraphics[width=0.9\linewidth]{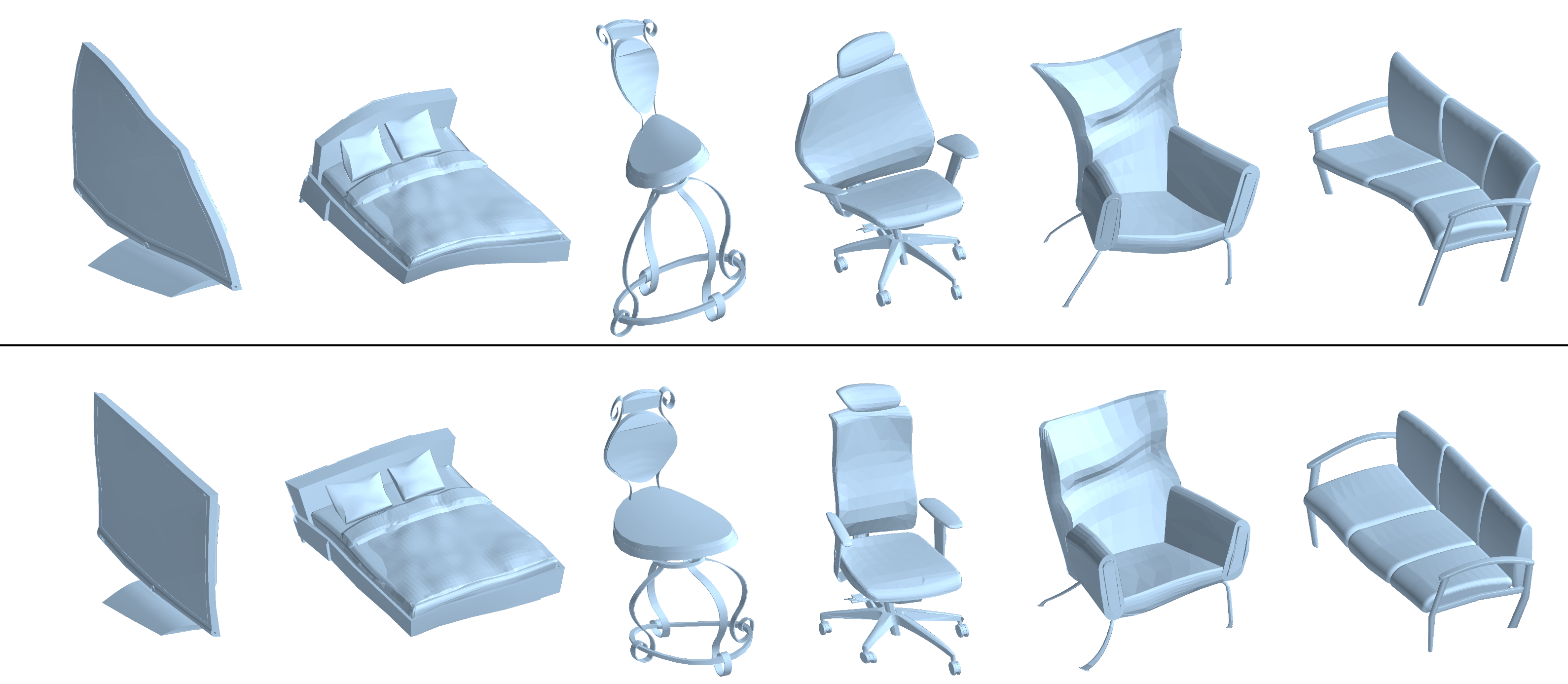}
\caption{3D shape rectification from highly non-linear deformations (top-row). Our method implicitly learns non-linear rectification transformations through a weakly supervised GAN process (bottom-row).}
\label{fig:tps_rect}
\end{figure*}

In Figure~\ref{fig:tps_rect} we show examples of non-linear shape rectification results obtained by training weakly-supervised a model coupled with the differential rendering module. The top and bottom rows depict the deformed and rectified shapes, respectively. Note that weak supervision in the form of simple discrimination loss between deformed and undeformed shapes was sufficient to learn highly non-linear 3D rectification transformations.

We non-linearly deform shapes using TPS defined by a $4\times 4$ control point grid. We randomly perturbate control points by adding noise drawn from a normal distribution with standard deviation $0.07$.
We quantitatively measure rectification performance by computing the correspondence distance between corresponding points in the deformed and rectified shape. We observe that RMSE of correspondence distance drops from initial 0.04 to 0.025 in final optimization iteration.

\begin{table}[]
	\centering
	\caption{Panoramic Rendering}
	\label{tab:panoramic_accuracy}
	\begin{tabular}{|l|l|}
		\hline
		\multicolumn{1}{|c|}{Method} & \multicolumn{1}{c|}{Accuracy}\\ \hline
		DeepPano (no clutter)        & 77.6 \\ \hline
		No adaptive rendering        & 72.2 \\ \hline 
		With adaptive rendering      & 78.2 \\ \hline 
	\end{tabular}
\end{table}

\begin{table}[]
	\centering
	\caption{Classification accuracy on randomly-rotated ModelNet40 for different methods}
	\label{tab:rot_class}
	\begin{tabular}{|l|l|l|l|l|}
		\hline
		\multicolumn{1}{|c|}{Method}                & \multicolumn{1}{|c|}{Representation}        & \multicolumn{1}{|c|}{Accuracy}  \\ \hline 
		Roveri et al.         & Point-Cloud/Depth-Map & 84.9             \\ \hline
		Roveri et al. PCA     & Point-Cloud/Depth-Map & 84.4             \\ \hline
		Roveri et al. Learned & Point-Cloud/Depth-Map & 85.4             \\ \hline
		PointNet              & Point-Cloud           & 85.5             \\ \hline
		Ours                  & Point-Cloud/Depth-Map & 87.4             \\ \hline
	\end{tabular}
\end{table}

\section{Conclusions and future work}
\label{sec:future}

In this paper we introduce a novel differential rendering module which defines a grid of cells in 3D space. The cells are functions which capture the 3D shape and their function parameters may be optimized in a gradient descend manner.
The compatibility of our differential rendering network with neural networks allows to optimize the rendering function and geometric transformations, as part of larger CNNs in an end-to-end manner.
This enables the models to implicitly learn 3D shape transformations and renderings in a weakly supervised process.
Results demonstrate the effectiveness of our method for processing 3D data efficiently.
\subsubsection{Limitations and future work.}
Our method essentially computes non-linear projections of the 3D data onto a 2D grid.
Our cells may fail to correctly render shapes consisting intricate geometries and topologies or with large hidden structures. For such shapes, it is impossible or extremely challenging to define a mapping between the 3D shape and a nonlinear 2D sheet in space. For such shapes, our differential rendering module seems inadequate (as well as any other rendering approach).

In the future, we plan to follow on this path and utilize differential rendering for additional weakly supervised 3D tasks such as shape completion and registration.
Furthermore, we plan to investigate utilization of our rendering module in the context of rotational invariant representations and processing.

\bibliography{paper}
\bibliographystyle{IEEEtran}

\end{document}